\documentclass[10pt,twocolumn,letterpaper]{article}

\usepackage[utf8]{inputenc}
\usepackage{cvpr}
\usepackage{times}
\usepackage{epsfig}
\usepackage{graphicx}
\usepackage{amsmath}
\usepackage{amssymb}
\usepackage[numbers]{natbib}
\usepackage{ctable}
\usepackage{framed,multirow}
\usepackage{placeins}

\newcommand{\quotes}[1]{``#1''}

\cvprfinalcopy
\def\cvprPaperID{31}

\ifcvprfinal\fi
\begin{document}

\title{Less is More: Sample Selection and Label Conditioning \\ Improve Skin Lesion Segmentation}

\author{Vinicius Ribeiro\textsuperscript{1} ~~~Sandra Avila\textsuperscript{2} ~~~Eduardo Valle\textsuperscript{1}\\
\textsuperscript{1}School of Electrical and Computing Engineering (FEEC) ~~~\textsuperscript{2}Institute of Computing (IC)\\
RECOD Lab., University of Campinas (UNICAMP), Brazil\\
}

\maketitle
\thispagestyle{empty}

\begin{abstract}

Segmenting skin lesions images is relevant both for itself and for assisting in lesion classification, but suffers from the challenge in obtaining annotated data. In this work, we show that segmentation may improve with less data, by selecting the training samples with best inter-annotator agreement, and conditioning the ground-truth masks to remove excessive detail. We perform an exhaustive experimental design considering several sources of variation, including three different test sets, two different deep-learning architectures, and several replications, for a total of 540 experimental runs. We found that sample selection and detail removal may have impacts corresponding, respectively, to 12\% and 16\% of the one obtained by picking a better deep-learning model.

\end{abstract}

\section{Introduction}\label{sec:intro}

Withholding data to improve machine learning is counter-intuitive, but we will show it brings promising improvements for skin lesion segmentation, by selecting the training samples with best inter-annotator agreement, and conditioning the ground-truth masks to remove excessive detail.

Segmenting skin lesions images, i.e., delimiting the lesion from the surrounding skin, is frequently employed both as an end-result and as an adjutant for lesion classification. Segmentation, however, is a very challenging task, in part due to the difficulty in obtaining properly annotated data.

All supervised machine-learning models need annotated data to be trained, posing unique challenges for medical tasks, where the annotators are specialists whose time is costly and scarce. Segmentation poses additional challenges since the annotations are intricate region borders instead of a single label. Researchers have attempted circumventing the need for data, with techniques like data augmentation~\citep{bisla2019towards, perez2018data}, the creation of artificial data from generative models~\citep{bissoto2018skin}, or even the use of self-supervision, which allows performing part of the training of supervised models without labels~\citep{jing2019self}.

\textit{Quality} of training data is another important dimension, in addition to \textit{quantity}. For skin lesion images, recent works have addressed that issue, exploring the lack of inter-annotator agreement in the ground truth of segmentation images \citep{ribeiro2019handling}, and the possibility of using annotations of different levels of confidence and granularity to learn segmentation masks~\citep{mirikharaji2019learning}.

In this work, we will follow a novel perspective: instead of finding ways to \textit{amplify} training data, we will show how less of it can enhance results. The two main contributions of this work are:

\begin{itemize}
  \item We show how training sample selection, based on inter-annotator agreement, can improve segmentation results, even when such selection is not applied to the test sets;
  \item We show how removing details from the ground-truth masks — using very simple \quotes{conditionings}  — can improve segmentation results, even when the same details are still required in the test set, i.e., when the test masks are not \quotes{conditioned}.
\end{itemize}

We evaluate those contributions thoroughly, in an exhaustive experimental design that considers several sources of variation, including three different test sets, two different deep-learning architectures, and several replications, for a total of 540 experimental runs. 

The remainder of the text is organized as follows. We survey the related works in \ref{sec:review}.
We present the sample selection technique in \ref{sec:selection}, and the ground-truth conditioning in \ref{sec:conditioning}. We provide details about our datasets, models, implementation, and experimental design in \ref{sec:materials}, with results following in \ref{sec:results}. We conclude the paper in \ref{sec:conclusions}.

\section{Literature Review}\label{sec:review}

Segmenting skin-lesion images has attracted scientific interest since the inception of automated skin-lesion analysis~\citep{dhawan1992segmentation, moss1996unsupervised}. Early lesion classification tended to mimic medical procedures~\citep{fornaciali2016towards}, such as the ABCD rule~\citep{nachbar1994abcd}, in which estimating, e.g.,  \textit{B}order irregularity and large \textit{D}iameter relied on segmentation. Such methods were also consonant with early computer vision art, in which segmentation was considered a crucial preliminary step for classification (e.g., to allow extracting shape features). \citet{celebi2015state} provide a comprehensive survey of early works on skin-lesion image segmentation.

We limit our analysis in this section to an overview of the field and promising methods proposed after deep learning. For a more comprehensive view, we reference a survey published by \citet{celebi2015state} that presents an overview of 50 published articles describing the state of the art of border detection algorithms. The survey reviews the pre-processing, segmentation methods, post-processing, and evaluation criteria of several works related to the area. It then presents a comparison of the methods concerning different aspects.

The transition of computer vision art to bags-of-words models in the 2000s \citep{sivic2006video}, and to deep learning in the 2010s \citep{krizhevsky2012imagenet} spelled the end of the viewpoint of segmentation as an ancillary technique in preparation for classification. That understanding, however, also increased the appreciation of segmentation for its own merits. With the accumulated experience brought by collective efforts like the PASCAL VOC \citep{everingham2010pascal} and ImageNet \citep{deng2009imagenet} challenges, we now understand not only that segmentation and classification can be tackled independently, but also that segmentation is usually \textit{much more challenging} than classification.

Those advances in computer vision appear in the current art in skin lesion analysis \citep{valle2020data, oliveira2018computational}, in which, although lesion segmentation is sometimes still used to help in the classification, it is largely understood as an important and challenging task in itself.

Deep learning underpins current art on skin-lesion segmentation. In this survey, we will highlight only a few works relevant to our discussion, and refer the reader to the reviews of \citet{tajbakhsh2020embracing} and \citet{kalinin2020medical}, on medical image segmentation, for a more broad survey of deep-learning-based techniques.

The ISIC Challenges of 2017~\citep{isic2017} and 2018~\citep{isic2018} included a segmentation task, and fostered several techniques. In 2017, a fully convolutional-deconvolutional network achieved 1st place~\citep{yuan2017automatic}, while the U-Net~\citep{ronneberger2015u} appeared in 2nd place~\citep{berseth2017isic}, and the ResNet~\citep{he2016deep} appeared in 3rd~\citep{bi2017automatic}. In 2018, a two-stage method based on MaskRCNN~\citep{he2017mask}, DeepLab~\citep{chen2017rethinking} and PSPNet~\citep{zhao2017pyramid} achieved 1st place~\citep{qian2018two}, while a simpler scheme with the DeepLab and transfer learning from VOC PASCAL 2012 achieved 2nd place~\citep{hao2018techreport}, and a traditional \quotes{U-Net-like} architecture, with ResNet-based encoder and decoder achieved 3rd place ~\citep{yuanfeng2018techreport}.

A recent development in skin-lesion segmentation is the use of generative models. \citet{xue2018adversarial} proposed SegAN, an end-to-end adversarial network architecture with multi-scale loss, and achieved 4th place at the 2018 ISIC Challenge.

Training, and especially, \textit{evaluating} machine-learning models, require accurate annotations. \citet{ribeiro2019handling} find, however, that information about inter-annotator agreement in visual datasets is very scarce, and when present, suggest a large variation among different tasks. In particular, for skin-lesion segmentation, they find the degree of agreement is only moderate, with a considerable portion of the samples having very poor inter-annotator agreements.

There are different solutions to that issue. On the one hand, we may ameliorate the quality of the annotations. Because reannotating the data is very expensive, \citet{ribeiro2019handling} suggest conditioning operations on the ground-truth masks that remove details, improving their agreement.

On the other hand, we may render our models less sensitive to noise. Deep learning models are, by nature, fairly insensitive to noisy annotations~\citep{rolnick2017deep}. An in-depth survey of segmentation techniques for medical images from noisy datasets~\citep{tajbakhsh2019embracing} addresses both the issue of scarce and imperfect annotations, and, for the latter, lists techniques to deal with weak labels (in the technical sense of weakly supervised learning), sparse labels (only part of the image is annotated), and noisy labels (labels with ambiguities and inaccuracies). Specifically for skin-lesion segmentation,  \citet{mirikharaji2019learning} address a continuum of annotations, ranging from fully detailed ground-truths until progressively weaker ones, by using polygons with fewer vertices, and ending with just a bounding box. They proposed a spatial-adaptive reweighting to treat clean and noisy pixel-level annotations in the loss function.

In this work, we propose a third alternative: removing the noisy samples from the dataset and, following~\citet{ribeiro2019handling}, removing excessive detail from the ground truths on the remaining samples. While the focus of Ribeiro et al. was improving the inter-annotator agreement on the \textit{dataset}, here we focus on the \textit{machine-learning models} and evaluate the impact of removing those details on them.

\section{Sample Selection based on Inter-Annotator Agreement}\label{sec:selection}

As mentioned, \citet{ribeiro2019handling} found a broad diversity in the inter-annotator agreement for the ISIC dataset images. In particular, the authors noticed a fairly \quotes{heavy tail} of very discordant annotations in their observation.

In this work, we evaluate the actual effect of those observations on segmentation models, by contrasting models learned the usual way, without any data selection, with models learned with fewer samples, eliminating the worst discordant samples in the tail.

To perform a fair comparison, we first selected all samples from the online ISIC Archive dataset with at least two segmentation ground-truth annotations. For each of those samples, we computed the average pairwise Cohen's Kappa score~\cite{mchugh2012interrater} for all existing ground-truth annotations. All samples with an average score above 0.5 went to the \textbf{best samples} dataset, and all samples, however the score, went to the \textbf{all samples} dataset.

Details about the data and selection procedure are in Sections~\ref{sec:dataset}, \ref{sec:models}, \ref{sec:analysis}  and~\ref{sec:results_selection}.

\section{Detail Elimination with Label Conditioning}\label{sec:conditioning}

In order to enhance inter-annotator agreement, \citet{ribeiro2019handling} propose applying \quotes{conditionings} on the ground-truth segmentation masks, which consist of eliminating details from them. They evaluate (in growing aggressiveness) the morphological operations of opening and closing, the convex hull, the morphological operations combined with the convex hull, and a bounding box. 

In this work, we follow up on the idea of conditioning the ground-truth masks, from a different point of view: the machine-learning model. Instead of measuring how much different conditionings affect the mask agreement, we will measure how they affect both the training and the evaluation (when applied to the test set) of segmentation models.

\begin{figure}
\begin{center}
\includegraphics[width=\linewidth]{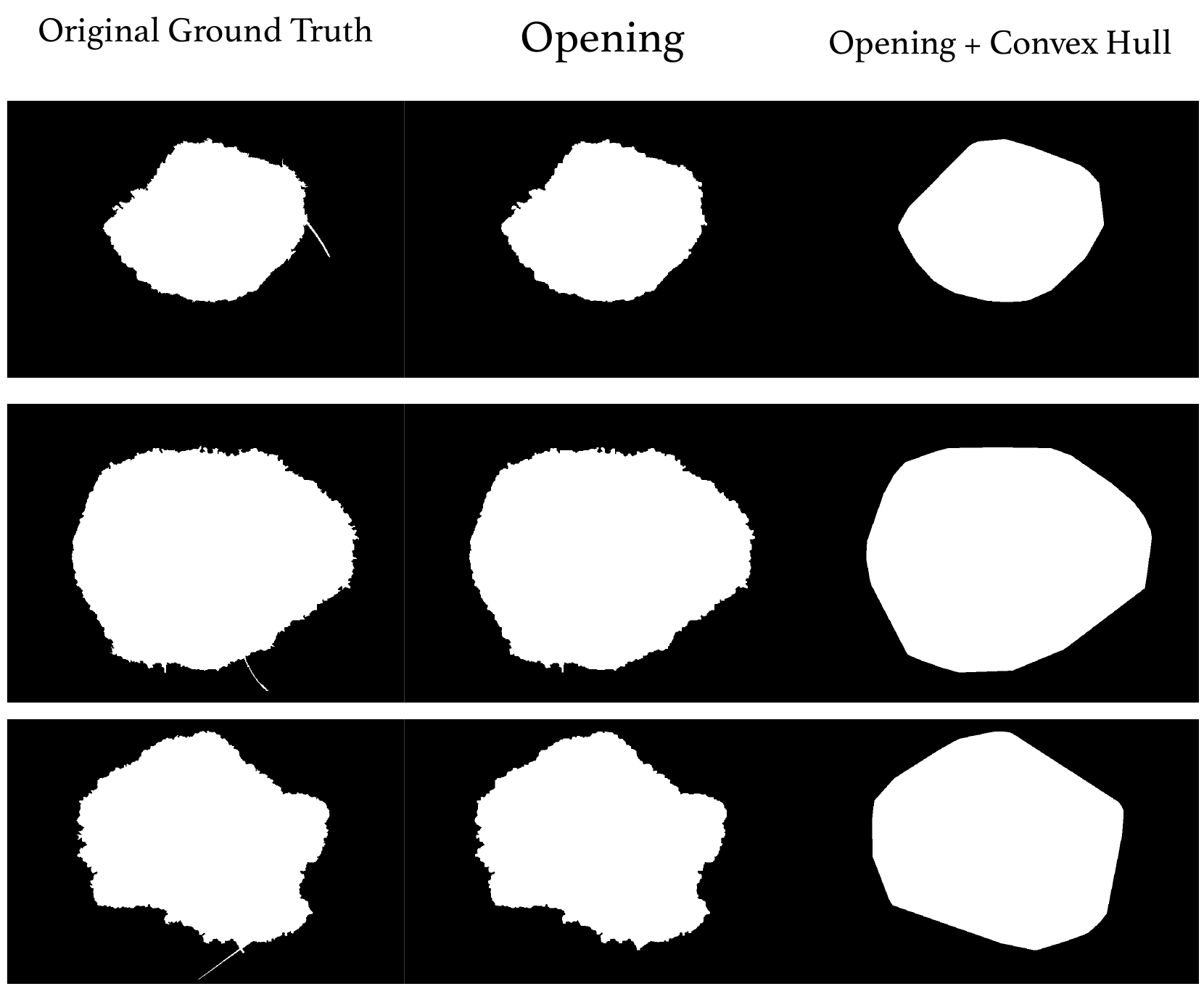}
\end{center}
   \caption{Three ground-truth segmentation masks from the ISIC Archive and the result of their conditioning with the two techniques we assessed. Conditioning removes the small details which may prove distracting for the models.}
\label{fig:transforms}
\end{figure}

In addition to the original images, we selected the two most promising conditionings proposed by \citet{ribeiro2019handling} for evaluation (\ref{fig:transforms}):
\begin{description}
  \item[None] no conditioning: the original images — used as a control;
\item[Opening] this morphological operation removes details like small protrusions in the lesion area. The structuring element was a 5-pixel-wide square;
  \item[Convex Hull] opening, just as above, followed by taking the convex hull, i.e., finding the tightest convex polygon that contains the lesion area.
\end{description}

We may interpret conditioning as denoising operations, aiming at preserving the cogent information while discarding details that arise from choosing a particular annotator. Our hypothesis, in this work, is that those annotator-dependent details may prove an expensive distraction for the models to learn.

Details about the procedure are in Sections~\ref{sec:models} and~\ref{sec:results_conditioning}.

\section{Materials and Methods}\label{sec:materials}

\subsection{Datasets}\label{sec:dataset}
\begin{figure}
\begin{center}
\includegraphics[width=0.9\linewidth]{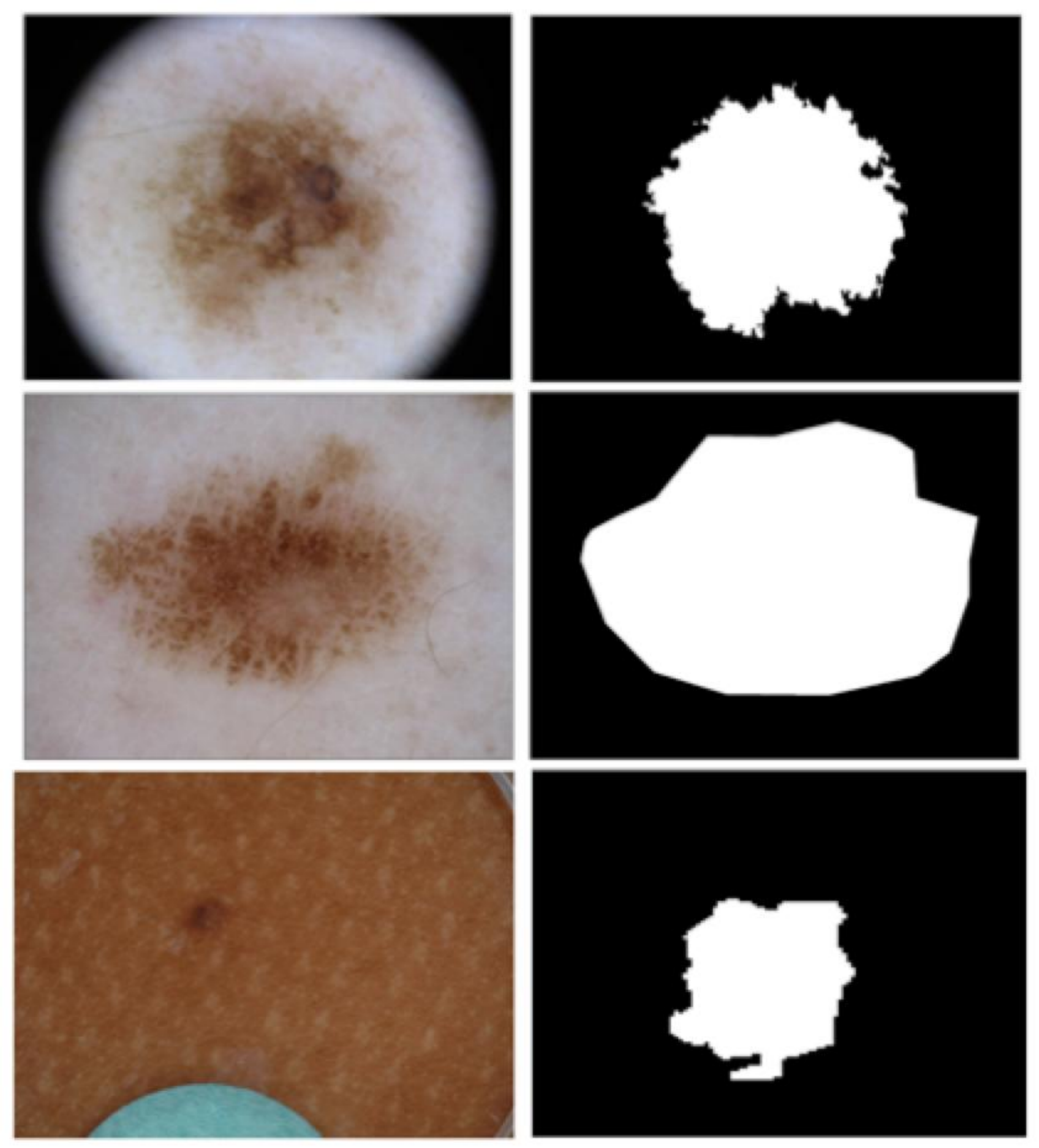}
\end{center}
   \caption{The three methods used to create the ISIC Archive segmentation masks. A flood-fill algorithm controlled by the annotator (top) tends to create very irregular borders; Manual polygon tracing (middle) creates very smooth borders; Fully-automated annotation validated by human annotator (bottom) is in-between, with borders that appear pixelated.}
\label{fig:high_variability}
\end{figure}

All training data used in this work came from the ISIC Archive~\citep{isicarchive} ---  curated by the International Skin Imaging Collaboration --- the largest publicly available dataset of images of skin lesions. Although a few other datasets also provide segmentation information \citep{ballerini2013color, mendonca2015ph2}, as far as we know, the ISIC Archive is the only public dataset with more than one segmentation annotation per lesion, and thus the only one where inter-annotator agreement can be appraised.

The ground truth annotations are highly variable due, in part, to three different methods to create the annotations (\ref{fig:high_variability}) and, in part, to differences of opinion and other specificities of human annotators.

At the time we collected our data, the ISIC Archive dataset contained 13\,779 images with segmentation ground truth masks, 2\,233 of those having multiple masks (\ref{table:annotations_distribution}). This latter number limited the training set for our experiments. We derived two training sets: one containing all 2\,233 samples (\textbf{all samples}), and other whose average pairwise Cohen's Kappa score between ground-truth masks was higher than $0.5$, (\textbf{best samples}). The latter had 1\,808 lesion images, i.e., only 81\% of the available samples.

We employed three datasets for testing the models. The first was formed by a random selection of 2\,000 images from the 11\,546 images of our ISIC Archive collection with only one segmentation mask. Two others are the PH2 dataset~\citep{mendonca2015ph2} collected at the Porto University, with 200 dermoscopic images, and the Edinburgh Dermofit Library~\citep{ballerini2013color}, with 1\,300 focal high-quality clinical images. Since the inter-annotator agreement in those datasets cannot be appraised, all three of them represent \quotes{in-the-wild} situations, without sample filtering. The first dataset represents the typical machine-learning evaluation pipeline, with training and evaluation being subsamples of the same dataset. In contrast, the two others represent a cross-dataset scenario, that challenges the generalization abilities of the models.

\begin{table}[tb]
\begin{center}
\begin{tabular}{rr}
\toprule
\# of Masks & \# of Samples \\
\midrule
1 & 11\,546 \\
2 & 2\,094 \\
3 & 100 \\
\textgreater4 & 39 \\
\midrule
Total & 13\,779 \\
\bottomrule
\end{tabular}
\end{center}
\caption{Distribution of samples by number of ground-truth segmentation masks in our ISIC Archive collection.}
\label{table:annotations_distribution}
\end{table}

\subsection{Models and conditionings}\label{sec:models}

LinkNet~\citep{chaurasia2017linknet} is a traditional \quotes{U-Net-like} architecture: encoder–decoder with skip connections between them. DeepLab V3+~\citep{chen2017rethinking}, in contrast, uses ResNet as the primary feature extractor, introduces new residual blocks for learning multi-scale features, and employs atrous convolutions with different dilation rates in the last residual block to better context understanding and scale invariance.

To train the networks, we split the training samples into 80/20 training and validation sets (\ref{table:full_clean_distributions}). All lesions have more than one ground-truth mask during training; we randomly select which mask to use every time we pick a sample to compose a batch. Thus, different masks may appear at different times during training. For model selection during validation, for each sample, we evaluate the target metric (Jaccard index) using all available annotations and retain the best (i.e., the highest). The test datasets have a single annotation per lesion, so mask selection and metric computation are straightforward.

We trained each model for 100 epochs, without early stopping, with an Adam~optimizer~\citep{kingma2014adam} and learning rate of 0.003. The loss function was a weighted sum of the soft Jaccard with the Binary Cross Entropy with Logits~\citep{iglovikov2017satellite}, with weights, respectively, of 8 and 1.

We applied three data augmentations, aiming to teach our model to be invariant to noise, color, and contrast. We add to each sample a Gaussian noise with zero mean and standard deviation of 2. We also add a color and a contrast enhancement, each parameterized by a Gaussian factor with mean 0.5 and standard deviation of 0.1, implemented using the Pillow image library \citep{lundh2012python}.

\begin{table}[tb]
    \begin{center}
        \begin{tabular}{rrr}
            \toprule
                  & \multicolumn{2}{c}{Training Set}\\
            Split & All Samples & Best Samples  \\
            \midrule
            Training & 1\,786 & 1\,449 \\
            Validation & 447 & 359 \\
            \midrule
            Total & 2\,233 & 1\,808 \\
            \toprule
        \end{tabular}
    \end{center}
    \caption{Training sets and their splits.}
    \label{table:full_clean_distributions}
\end{table}

We implemented the models, training, and evaluation pipelines using the PyTorch framework for deep learning~\citep{paszke2017automatic}. 
We developed all the conditionings in Python, using the morphology package of the scikit-image library~\citep{van2014scikit}, and auxiliary code in NumPy~\citep{oliphant2006guide}. During validation, we always apply to the masks the same conditioning used for training.

All code necessary to reproduce this work is available at our Github repository\footnote{https://github.com/vribeiro1/skin-lesion-segmentation-agreement}.

\subsection{Experimental design}\label{sec:design}

We ran a single exhaustive experimental design to validate both the sample selection and the ground-truth conditioning. The design also aimed at capturing sources of variation present in the actual deployment of segmentation models and included the following factors:

\begin{description}
\item[Training set] This can be either \textbf{all samples} of our ISIC collection, or a selection of the \textbf{best samples}, whose ground-truth segmentation masks have an average pairwise Cohen's Kappa agreement above 0.5 (details in \ref{sec:dataset}).
\item[Test set] A split from our \textbf{ISIC} subset (no sample selection), the \textbf{PH2} dataset, or the \textbf{Dermofit} dataset. The latter two are a cross-dataset evaluation (details in \ref{sec:dataset}).
\item[Training conditioning] Conditioning applied on the ground-truth of the samples used for training the model. \textbf{None} for the original image, \textbf{opening} for the morphological operator removing small details, and \textbf{convex hull} for opening followed by taking the convex hull (details in \ref{sec:conditioning}).
\item[Test conditioning] Conditioning applied on the ground-truth of the samples used for evaluating the model. The levels are the same as above.
\item[Model] One of two deep-learning models: \textbf{LinkNet} or \textbf{DeepLab} (details in \ref{sec:models}).
\end{description}

Each treatment was replicated 5 times for a total of 540 runs. In all experiments, the outcome was the segmentation accuracy, measured by the Jaccard index (sometimes named Intersection over Union or IoU). The Jaccard index is vastly employed in the semantic segmentation literature and it is the primary metric of the segmentation task in the 2017 editions of the ISIC Challenge.

The statistical analysis was a full factorial analysis of variance (ANOVA), which we used both to measure significance (p-values) and effect sizes ($\eta^2$). In addition to the statistical test, we employed interaction plots to elucidate the relationship between the factors.

\section{Results}\label{sec:results}

\begin{figure*}[tb]
\begin{center}
\includegraphics[width=0.48\textwidth]{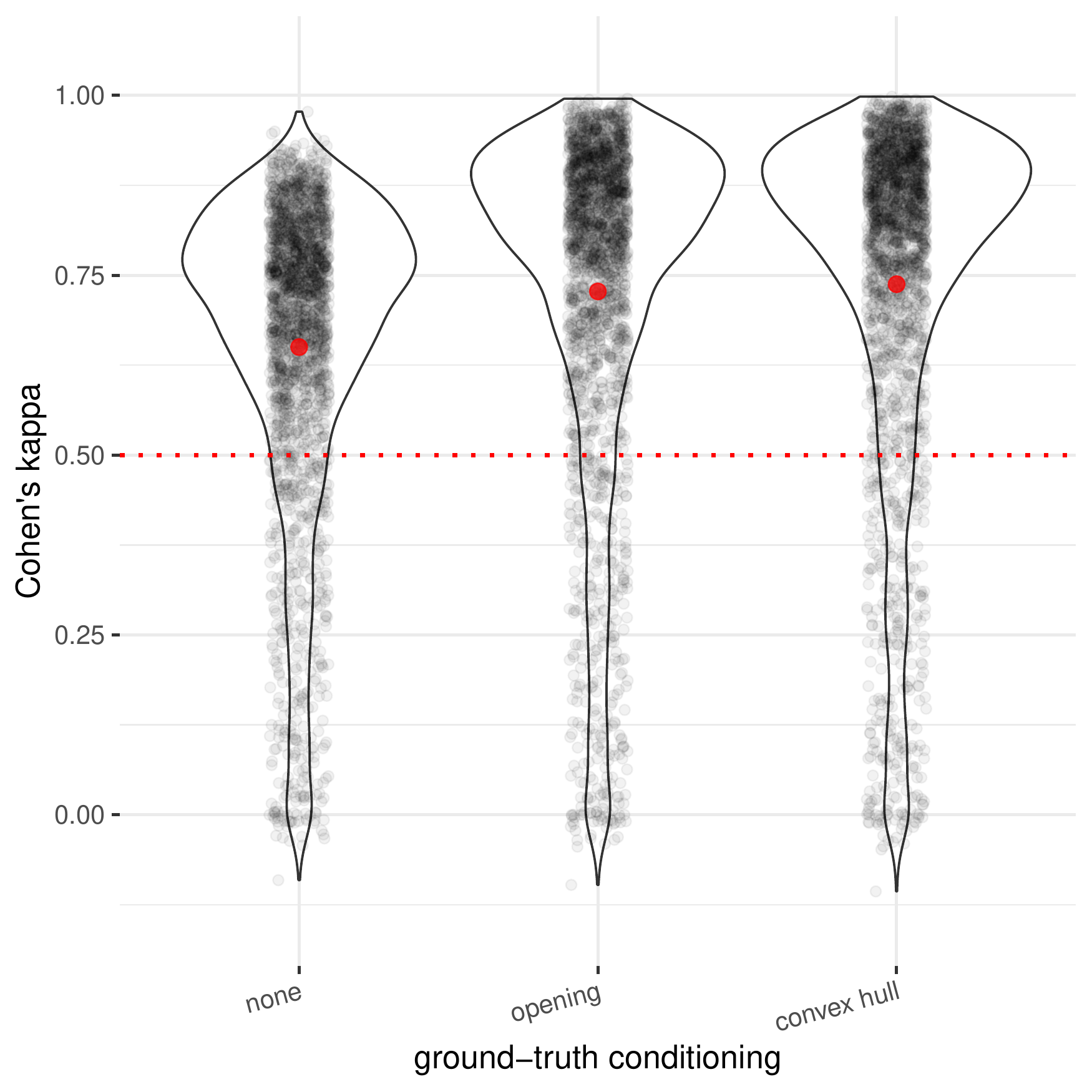}
\includegraphics[width=0.48\textwidth]{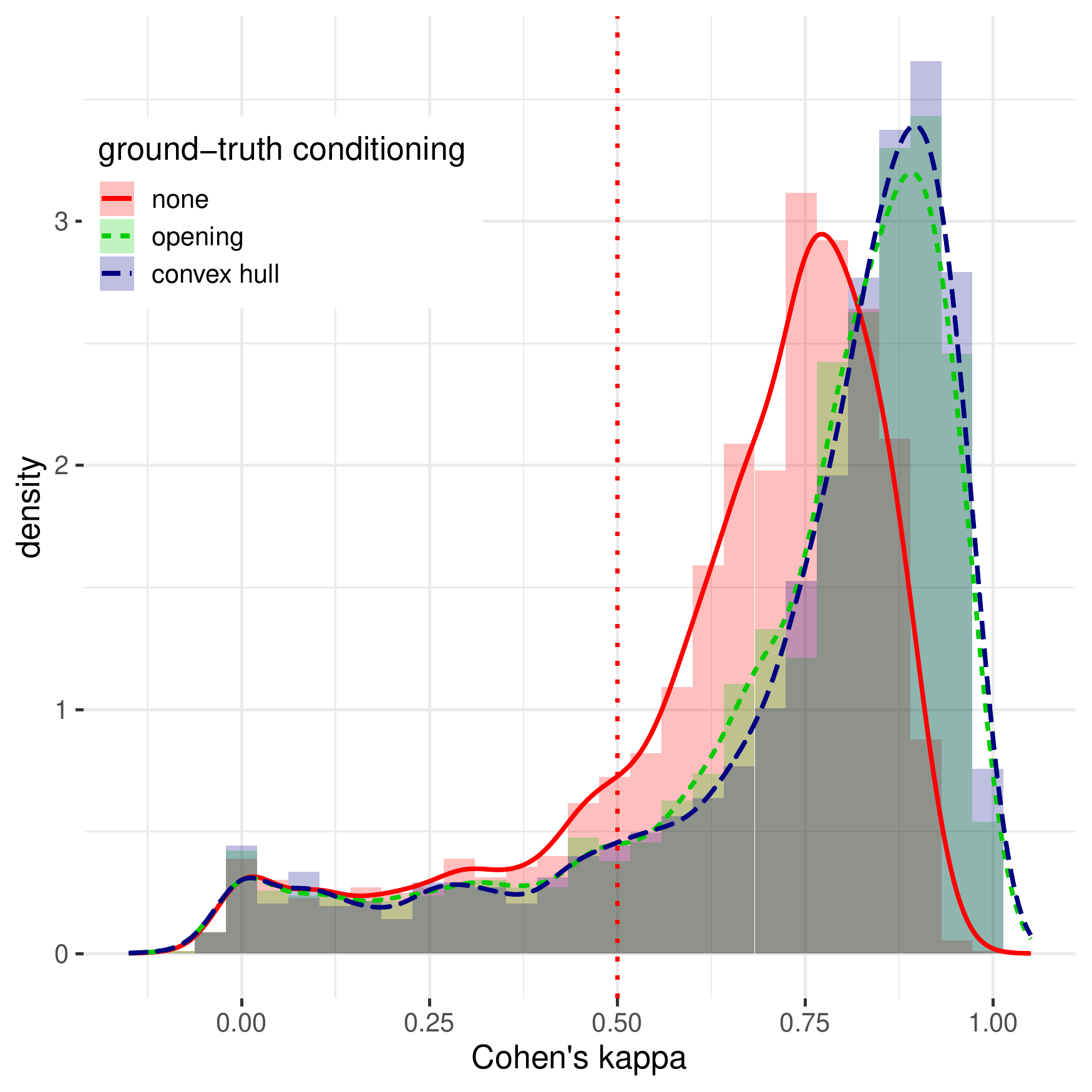}
\end{center}
  \caption{Distributions of inter-annotator agreements for the ground-truths pre- (none) and post- the evaluated conditionings (opening and convex hull) on the 2\,233 samples of the \textit{all samples} subset from our ISIC Archive collection. Both plots show the same data. Right: the histograms (shaded areas) and estimated densities (superimposed lines). Left: the original samples (black dots), the estimated densities (violin plots), and the estimated means (red dot). The red dotted line shows the threshold used for the \textit{best samples} dataset.\vspace{0.5cm}}
\label{fig:all_samples_kappa_distributions}
\end{figure*}

\begin{table}[tb]
\begin{center}
\begin{tabular}{rccccc}
\toprule
             & \multicolumn{5}{c}{Percentile} \\
Conditioning & 5 & 25 & 50 & 75 & 100 \\
\midrule
       none & 0.12 & 0.57 & 0.72 & 0.80 & 0.88  \\
    opening & 0.13 & 0.66 & 0.81 & 0.90 & 0.96  \\
convex hull & 0.13 & 0.67 & 0.83 & 0.90 & 0.96  \\
\bottomrule
\end{tabular}
\end{center}
\caption{Percentiles of the average pairwise Cohen's kappa score for the \textit{all samples} subset of our ISIC Archive collection.}
\label{table:all_samples_kappa_percentiles}
\end{table}

\subsection{Dataset analysis and sample selection}\label{sec:analysis}

\ref{fig:all_samples_kappa_distributions} shows the distribution of inter-annotator agreement, measured as the average pairwise Cohen's kappa score of the ground-truth masks, found in the 2\,233 samples of the \textit{all samples} subset from our ISIC Archive collection. The plots show both the original data (\textit{none} conditioning) and the data after the application of the \textit{opening} and \textit{convex hull} conditionings.

The improvement in agreement brought by the conditionings is visible as both the mode and the mean of the distributions are pushed towards higher kappa values. The conditionings are not, however, able to deal with large discordances in the annotations, and all distributions have a fairly heavy tail of very low kappa values. The percentiles of the kappa values in \ref{table:all_samples_kappa_percentiles} also reinforce those findings.

The dotted red line shows the threshold of 0.5 used to select the 1\,808 samples of the \textit{best samples} subset. Notice that this set has a fixed size since the selection is made on the unconditioned kappa values, regardless of the conditionings used in the experiment.

\subsection{Impact of sample selection}\label{sec:results_selection}

\ref{fig:all_samples_kappa_distributions} is an interaction plot highlighting the effect of sample selection, the choice between the \textit{all samples} vs. \textit{best samples} in the \textit{training set} factor of our experimental design. The average effect of that choice can be appreciated on the solid lines in that plot, where selecting the best samples for training appears systematically above picking all samples. Recall that this implies discarding almost 20\% of the training samples, and that such selection is \textit{not} performed on the test sets. Those results are far from trivial, since deep-learning if fairly robust to noise \citep{rolnick2017deep} and often presents better results in larger noisier dataset than in smaller cleaner ones.

In addition to those averaged aggregate results, two other results deserve attention. First, there are interactions among sample selection on the training set, conditioning on the training set, and conditioning on the test set, with those facts act synergetically. We will explore those interactions in more detail in \ref{sec:anova}. Second, the results may vary according to the test set. Indeed, for the PH2 dataset there is a slight inversion of the results (although the experiments are quite mixed, as the individual data points show). In contrast, the Edinburgh Dermofit dataset shows the largest positive differences, which is remarkable given that dataset has focal clinical images instead of dermoscopic images and, thus, poses the widest generalization gap for the models to bridge.

\begin{figure*}[tb]
\begin{center}
\includegraphics[width=0.99\textwidth]{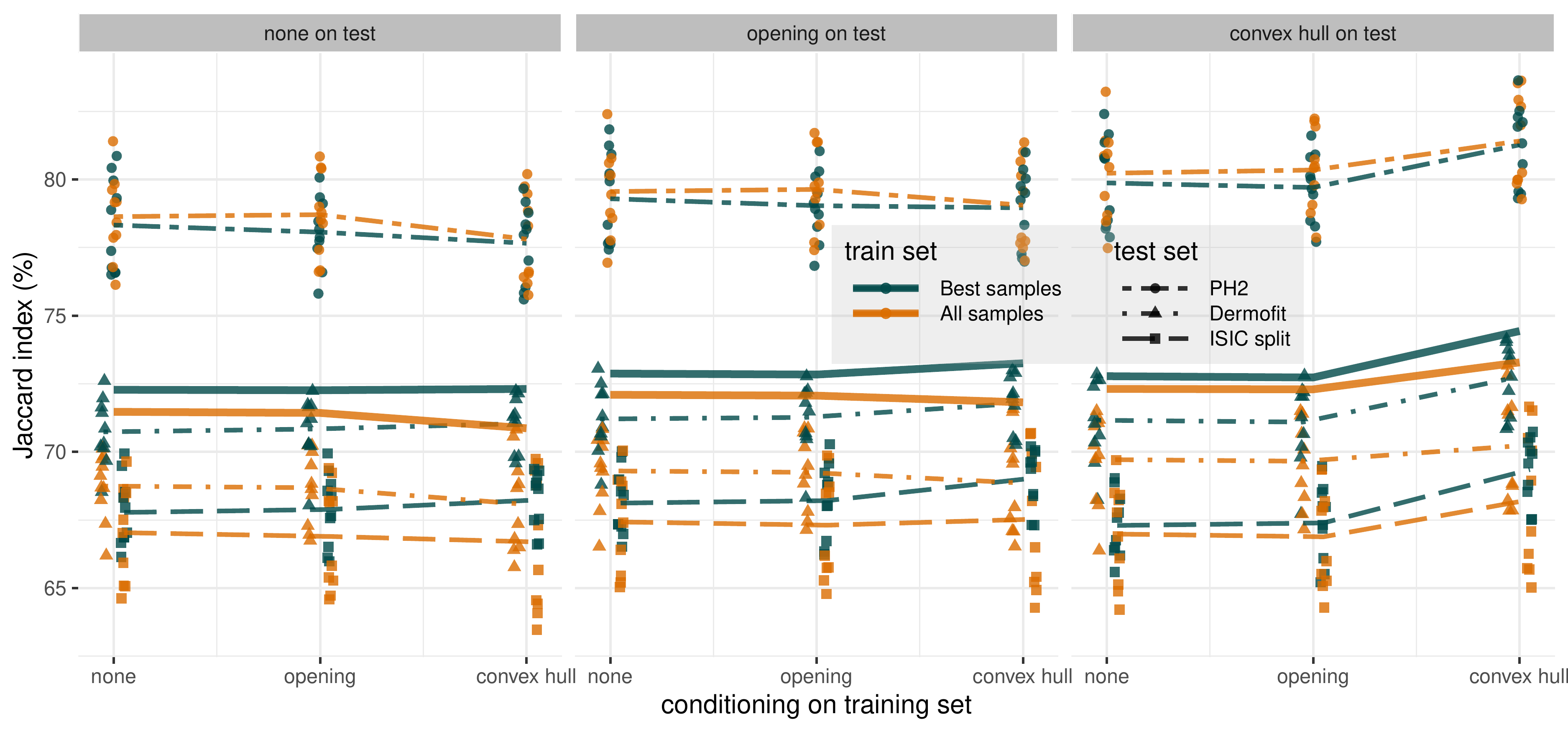}
\end{center}
   \caption{Interaction plot highlighting the effect of sample selection, i.e., the \textit{training set} factor in our experimental design. Points are individual experiments, dashed lines are averages per test set, solid line is the average for all three test sets. Each panel show the results for a choice of conditioning on the test set (leftmost: no conditioning). Sample selection improves results across a large variation of treatments.}
\label{fig:sample_selection}
\end{figure*}

\begin{figure*}[tb]
\begin{center}
\includegraphics[width=0.99\textwidth]{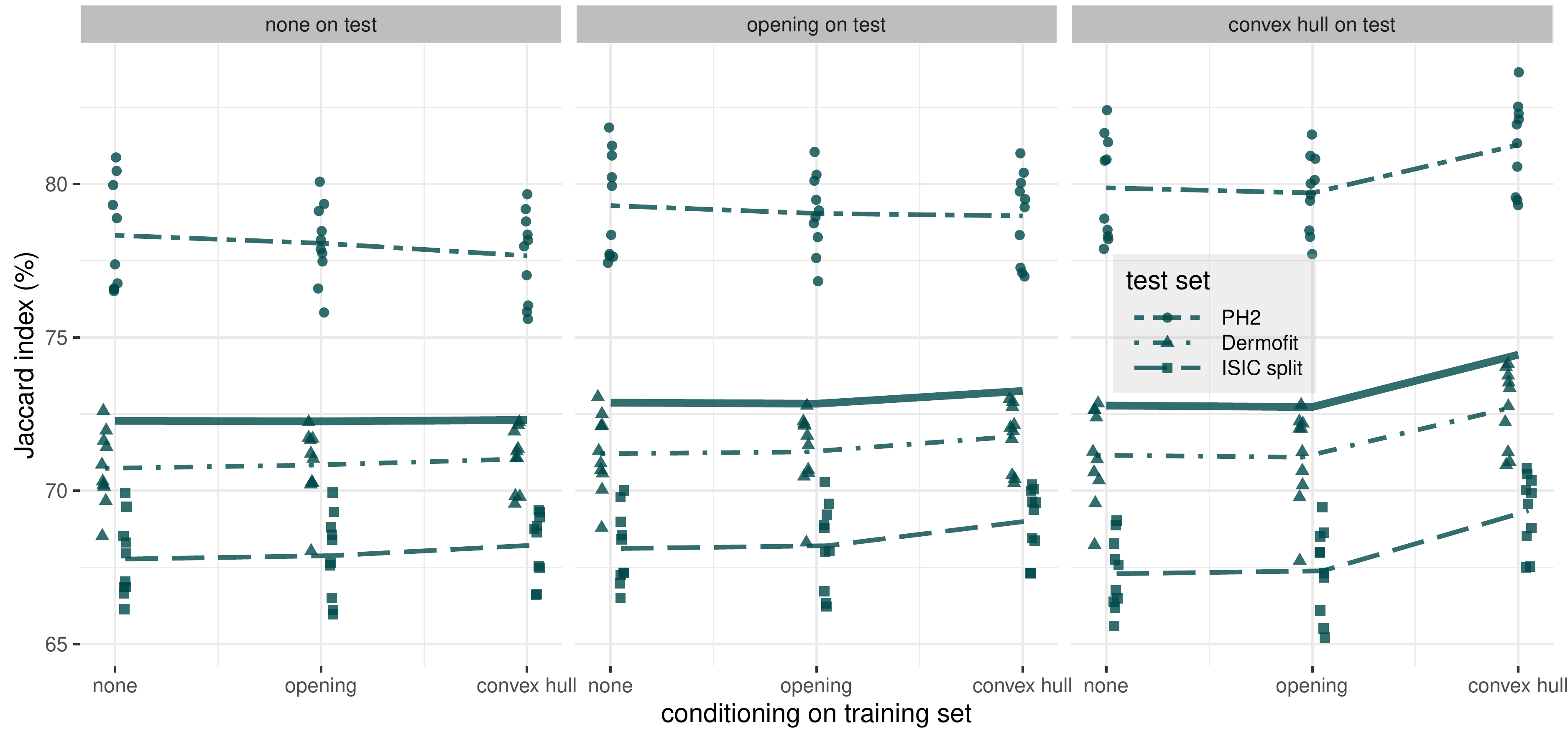}
\end{center}
   \caption{Same interactions as shown in \ref{fig:sample_selection}, but plotting only experiments trained with \textit{best samples}. Picking just one of the training sets highlights the two conditionings due to the interaction we found between the three factors.}
\label{fig:conditioning}
\end{figure*}

\subsection{Impact of ground truth conditioning}\label{sec:results_conditioning}

\ref{fig:conditioning} is an interaction plot highlighting the effect of conditioning the ground-truth masks on the training and the test sets. The plot shows only the results for the \textit{best samples} dataset because we found important positive interactions between conditioning and sample selections.

The most surprising result is that removing details from the ground truths on the training set does not reduce the performance, on average, of models — even when those same details are required on the test sets (leftmost panel). The exact results varied by test set, with PH2 showing a slight decrease, and ISIC and Edinburgh Dermofit showing a slight \textit{increase} in performance. Those results showcase that adding excessive detail in the ground-truth masks may be counterproductive.

That is particularly true when those details are not needed for the target application. The rightmost panel shows that when the convex hull conditioning is applied to both training and test ground truth masks — to evaluate, e.g., an application where the rough contour of the lesion is enough — the results sharply increase, for all three test sets.

\subsection{Statistical analysis}\label{sec:anova}

We can divide, in our statistical analysis, the sources of variation in three groups:

\begin{description}
\item[Design factors] Those are the factors we could actively control in the actual deployment of a machine-learning model. In our experiment, those are \textit{model}, \textit{training set}, \textit{training conditioning}, and \textit{test conditioning}.
\item[Nuisance factors] Those are the factors we cannot control in any actual deployment of a model, but we can control in an experiment. In our experiment, the single factor in this category is \textit{test set}.
\item[Uncontrolled sources] Those are sources of variation we cannot control in either situation: fluctuations in training (random seeds, numerical errors, etc.), hardware fluctuations, etc.
\end{description}

The statistical analysis was a full factorial ANOVA. All factors were found significant, with tiny p-values ($\sim10^{-6}$ for test conditioning, $\sim10^{-16}$ for all others). We considered up to 3rd order interactions, and several of them were significant, notably almost all 2nd order interactions (the exceptions were \textit{training set} with \textit{test conditioning} and \textit{model} with \textit{test conditioning}). 

The main source of variation was \textit{test set}, which explained 88\% of the global variation (i.e., $\eta^2=0.88$ effect size). That is perhaps unsurprising considering the three datasets varied widely in difficulty, with the PH2 dataset being much easier to segment than the other two. Uncontrolled sources accounted for less than 2\% of the variation. The remainder variation was scattered among the other factors and interactions, \textit{model} being the largest by far (6\% of the variation).

Considering only the variation we can design for, \textit{model} was the most influential, explaining almost 57\% of it. Sample selection was considerable, with \textit{training set} gathering over 7\% of the variation. \textit{Training conditioning} alone explained just a little over 1\% of the variation, while \textit{test conditioning} alone gathered almost 9\%. In addition, both factors interacted to explain almost 4\% of the \quotes{designable} variation.  

\section{Conclusions}\label{sec:conclusions}

As previously observed by \citet{ribeiro2019handling}, segmentation ground-truths for skin lesion images present substantial inter-annotator disagreement. Although that, for the moment, can only be measured on the ISIC Archive, there is no reason to believe the results would be different for other datasets, if they had more than one annotation available per sample. In this work, we showcased how a strategy of selecting the samples with largest disagreement may result in significantly improved performance. We also showed how removing details on the segmentation masks (by conditioning them with simple operators) may improve the results, especially if those details are not needed on prediction time.

To put our findings in perspective, consider the improvement brought from moving from LinkNet to DeepLab: this was the most important \quotes{designable} factor we found but, of course, creating new deep learning architectures is a laborious and haphazardous enterprise. One can obtain 12\% of that improvement simply by \textit{throwing away} 1/5th of the training data. By giving up detail on the segmentation masks, one can obtain 16\% of that improvement. For many applications of lesions segmentation (e.g., finding a rough contour, or determining a lesion diameter) less is more, in a very concrete sense.

The sample selection technique proposed in this paper requires multiple annotations per sample, a condition that makes it applicable to very few of the available training data. In the future, we would like to extend it to samples with a single ground-truth mask, greatly increasing its applicability.

\section{Acknowledgements}\label{sec:acks}

E. Valle is partially funded by a CNPq PQ-2 grant (311905/2017-0), and by a FAPESP grant (2019/05018-1). S. Avila is partially funded by Google Research Awards for Latin America 2018 \& 2019, FAPESP (2017/16246-0) and FAEPEX (3125/17). This project is partially funded by CNPq Universal grant (424958/2016-3). RECOD Lab. is partially supported by diverse projects and grants from FAPESP, CNPq, and CAPES. The funding sources had no involvement in the data acquisition, study design, result analysis, or in the manuscript writing. We gratefully acknowledge the donation of GPU Hardware by NVIDIA Corporation, used in this work.

{
\small
\bibliographystyle{plainnat}
\bibliography{egbib}

\begin{thebibliography}{46}
\providecommand{\natexlab}[1]{#1}
\providecommand{\url}[1]{\texttt{#1}}
\expandafter\ifx\csname urlstyle\endcsname\relax
  \providecommand{\doi}[1]{doi: #1}\else
  \providecommand{\doi}{doi: \begingroup \urlstyle{rm}\Url}\fi

\bibitem[isi()]{isicarchive}
{International Skin Imaging Collaboration: Melanoma Project}.
\newblock \url{https://isic-archive.com}.

\bibitem[Ballerini et~al.(2013)Ballerini, Fisher, Aldridge, and
  Rees]{ballerini2013color}
L.~Ballerini, R.~B Fisher, B.~Aldridge, and J.~Rees.
\newblock A color and texture based hierarchical k-nn approach to the
  classification of non-melanoma skin lesions.
\newblock In \emph{Color Medical Image Analysis}, pages 63--86. Springer, 2013.

\bibitem[Berseth(2017)]{berseth2017isic}
M.~Berseth.
\newblock Isic 2017-skin lesion analysis towards melanoma detection.
\newblock \emph{arXiv preprint arXiv:1703.00523}, 2017.

\bibitem[Bi et~al.(2017)Bi, Kim, Ahn, and Feng]{bi2017automatic}
L.~Bi, J.~Kim, E.~Ahn, and D.~Feng.
\newblock Automatic skin lesion analysis using large-scale dermoscopy images
  and deep residual networks.
\newblock \emph{arXiv preprint arXiv:1703.04197}, 2017.

\bibitem[Bisla et~al.(2019)Bisla, Choromanska, Berman, Stein, and
  Polsky]{bisla2019towards}
D.~Bisla, A.~Choromanska, R.~S. Berman, J.~A. Stein, and D.~Polsky.
\newblock Towards automated melanoma detection with deep learning: Data
  purification and augmentation.
\newblock In \emph{IEEE Conference on Computer Vision and Pattern Recognition
  Workshops}, 2019.

\bibitem[Bissoto et~al.(2018)Bissoto, Perez, Valle, and Avila]{bissoto2018skin}
A.~Bissoto, F.~Perez, E.~Valle, and S.~Avila.
\newblock Skin lesion synthesis with generative adversarial networks.
\newblock In \emph{OR 2.0 Context-Aware Operating Theaters, Computer Assisted
  Robotic Endoscopy, Clinical Image-Based Procedures, and Skin Image Analysis},
  pages 294--302. 2018.

\bibitem[Celebi et~al.(2015)Celebi, Wen, Iyatomi, Shimizu, Zhou, and
  Schaefer]{celebi2015state}
M.~E. Celebi, Q.~Wen, H.~Iyatomi, K.~Shimizu, H.~Zhou, and G.~Schaefer.
\newblock A state-of-the-art survey on lesion border detection in dermoscopy
  images.
\newblock \emph{Dermoscopy Image Analysis}, pages 97--129, 2015.

\bibitem[Chaurasia and Culurciello(2017)]{chaurasia2017linknet}
A.~Chaurasia and E.~Culurciello.
\newblock Linknet: Exploiting encoder representations for efficient semantic
  segmentation.
\newblock In \emph{2017 IEEE Visual Communications and Image Processing
  (VCIP)}, pages 1--4. IEEE, 2017.

\bibitem[Chen et~al.(2017)Chen, Papandreou, Schroff, and
  Adam]{chen2017rethinking}
L.C. Chen, G.~Papandreou, F.~Schroff, and H.~Adam.
\newblock Rethinking atrous convolution for semantic image segmentation.
\newblock \emph{arXiv preprint arXiv:1706.05587}, 2017.

\bibitem[Codella et~al.(2018)Codella, Gutman, Celebi, Helba, Marchetti, Dusza,
  Kalloo, Liopyris, Mishra, Kittler, et~al.]{isic2017}
N.~C.~F. Codella, D.~Gutman, M.~E. Celebi, B.~Helba, M.~A. Marchetti, S.~W.
  Dusza, A.~Kalloo, K.~Liopyris, N.~Mishra, H.~Kittler, et~al.
\newblock {Skin lesion analysis toward melanoma detection: A challenge at the
  2017 International Symposium on Biomedical Imaging (ISBI), hosted by the
  International Skin Imaging Collaboration (ISIC)}.
\newblock In \emph{IEEE International Symposium on Biomedical Imaging}, pages
  168--172, 2018.

\bibitem[Codella et~al.(2019)Codella, Rotemberg, Tschandl, Celebi, Dusza,
  Gutman, Helba, Kalloo, Liopyris, Marchetti, et~al.]{isic2018}
N.~C.~F. Codella, V.~Rotemberg, P.~Tschandl, M.~E. Celebi, S.~Dusza, D.~Gutman,
  B.~Helba, A.~Kalloo, K.~Liopyris, M.~Marchetti, et~al.
\newblock {Skin Lesion Analysis Toward Melanoma Detection 2018: A Challenge
  Hosted by the International Skin Imaging Collaboration (ISIC)}.
\newblock \emph{arXiv preprint arXiv:1902.03368}, 2019.

\bibitem[Deng et~al.(2009)Deng, Dong, Socher, Li, Li, and
  Fei-Fei]{deng2009imagenet}
J.~Deng, W.~Dong, R.~Socher, L.-J. Li, K.~Li, and L.~Fei-Fei.
\newblock Imagenet: A large-scale hierarchical image database.
\newblock In \emph{IEEE Conference on Computer Vision and Pattern Recognition},
  pages 248--255, 2009.

\bibitem[Dhawan and Sim(1992)]{dhawan1992segmentation}
A.~P. Dhawan and A.~Sim.
\newblock Segmentation of images of skin lesions using color and texture
  information of surface pigmentation.
\newblock \emph{Computerized Medical Imaging and Graphics}, 16\penalty0
  (3):\penalty0 163--177, 1992.

\bibitem[Du et~al.(2018)Du, Seok, Ng, Yuan, and Feng]{hao2018techreport}
H.~Du, J.~Young Seok, D.~Ng, N.~K. Yuan, and M.~Feng.
\newblock Team holidayburned at isic challenge 2018.
\newblock Technical report, 2018.

\bibitem[Everingham et~al.(2010)Everingham, Van~Gool, Williams, Winn, and
  Zisserman]{everingham2010pascal}
M.~Everingham, L.~Van~Gool, C.~K.~I. Williams, J.~Winn, and A.~Zisserman.
\newblock The pascal visual object classes (voc) challenge.
\newblock \emph{International Journal of Computer Vision}, 88\penalty0
  (2):\penalty0 303--338, 2010.

\bibitem[Fornaciali et~al.(2016)Fornaciali, Carvalho, Bittencourt, Avila, and
  Valle]{fornaciali2016towards}
M.~Fornaciali, M.~Carvalho, F.~V. Bittencourt, S.~Avila, and E.~Valle.
\newblock Towards automated melanoma screening: Proper computer vision \&
  reliable results.
\newblock \emph{arXiv preprint arXiv:1604.04024}, 2016.

\bibitem[He et~al.(2016)He, Zhang, Ren, and Sun]{he2016deep}
K.~He, X.~Zhang, S.~Ren, and J.~Sun.
\newblock Deep residual learning for image recognition.
\newblock In \emph{IEEE conference on computer vision and pattern recognition},
  pages 770--778, 2016.

\bibitem[He et~al.(2017)He, Gkioxari, Doll{\'a}r, and Girshick]{he2017mask}
K.~He, G.~Gkioxari, P.~Doll{\'a}r, and R.~Girshick.
\newblock Mask r-cnn.
\newblock In \emph{IEEE international conference on computer vision}, pages
  2961--2969, 2017.

\bibitem[Iglovikov et~al.(2017)Iglovikov, Mushinskiy, and
  Osin]{iglovikov2017satellite}
V.~Iglovikov, S.~Mushinskiy, and V.~Osin.
\newblock Satellite imagery feature detection using deep convolutional neural
  network: A kaggle competition.
\newblock \emph{arXiv preprint arXiv:1706.06169}, 2017.

\bibitem[Ji et~al.(2018)Ji, Li, Zhang, Lin, and Chen]{yuanfeng2018techreport}
Y.~Ji, X.~Li, G.~Zhang, D.~Lin, and H.~Chen.
\newblock Automatic skin lesion segmentation by feature aggregation
  convolutional neural network.
\newblock Technical report, 2018.

\bibitem[Jing and Tian(2019)]{jing2019self}
L.~Jing and Y.~Tian.
\newblock Self-supervised visual feature learning with deep neural networks: A
  survey.
\newblock \emph{arXiv preprint arXiv:1902.06162}, 2019.

\bibitem[Kalinin et~al.(2020)Kalinin, Iglovikov, Rakhlin, and
  Shvets]{kalinin2020medical}
Alexandr~A Kalinin, Vladimir~I Iglovikov, Alexander Rakhlin, and Alexey~A
  Shvets.
\newblock Medical image segmentation using deep neural networks with
  pre-trained encoders.
\newblock In \emph{Deep Learning Applications}, pages 39--52. Springer, 2020.

\bibitem[Kingma and Ba(2014)]{kingma2014adam}
D.~P. Kingma and J.~Ba.
\newblock Adam: A method for stochastic optimization.
\newblock \emph{arXiv preprint arXiv:1412.6980}, 2014.

\bibitem[Krizhevsky et~al.(2012)Krizhevsky, Sutskever, and
  Hinton]{krizhevsky2012imagenet}
A.~Krizhevsky, I.~Sutskever, and G.~E. Hinton.
\newblock Imagenet classification with deep convolutional neural networks.
\newblock In \emph{Advances in Neural Information Processing Systems}, pages
  1097--1105, 2012.

\bibitem[Lundh et~al.(2012)Lundh, Ellis, et~al.]{lundh2012python}
F.~Lundh, M.~Ellis, et~al.
\newblock Python imaging library (pil), 2012.

\bibitem[McHugh(2012)]{mchugh2012interrater}
M.~L. McHugh.
\newblock Interrater reliability: the kappa statistic.
\newblock \emph{Biochemia medica: Biochemia medica}, 22\penalty0 (3):\penalty0
  276--282, 2012.

\bibitem[Mendonca et~al.(2015)Mendonca, Celebi, Mendonca, and
  Marques]{mendonca2015ph2}
T.~F. Mendonca, M.~E. Celebi, T.~Mendonca, and J.~S. Marques.
\newblock Ph2: A public database for the analysis of dermoscopic images.
\newblock \emph{Dermoscopy image analysis}, 2015.

\bibitem[Mirikharaji et~al.(2019)Mirikharaji, Yan, and
  Hamarneh]{mirikharaji2019learning}
Z.~Mirikharaji, Y.~Yan, and G.~Hamarneh.
\newblock Learning to segment skin lesions from noisy annotations.
\newblock In \emph{Domain Adaptation and Representation Transfer and Medical
  Image Learning with Less Labels and Imperfect Data}, pages 207--215.
  Springer, 2019.

\bibitem[Moss et~al.(1996)Moss, Hance, Umbaugh, and
  Stoecker]{moss1996unsupervised}
R.~H. Moss, G.~A. Hance, S.~E. Umbaugh, and W.~V. Stoecker.
\newblock Unsupervised color image segmentation: with application to skin tumor
  borders.
\newblock 1996.

\bibitem[Nachbar et~al.(1994)Nachbar, Stolz, Merkle, Cognetta, Vogt,
  Landthaler, Bilek, Braun-Falco, and Plewig]{nachbar1994abcd}
F.~Nachbar, W.~Stolz, T.~Merkle, A.~B. Cognetta, T.~Vogt, M.~Landthaler,
  P.~Bilek, O.~Braun-Falco, and G.~Plewig.
\newblock The abcd rule of dermatoscopy: high prospective value in the
  diagnosis of doubtful melanocytic skin lesions.
\newblock \emph{Journal of the American Academy of Dermatology}, 30\penalty0
  (4):\penalty0 551--559, 1994.

\bibitem[Oliphant(2006)]{oliphant2006guide}
T.~E. Oliphant.
\newblock \emph{A guide to NumPy}, volume~1.
\newblock Trelgol Publishing USA, 2006.

\bibitem[Oliveira et~al.(2018)Oliveira, Papa, Pereira, and
  Tavares]{oliveira2018computational}
R.~B. Oliveira, J.~P. Papa, A.~S. Pereira, and J.~M. R.~S. Tavares.
\newblock Computational methods for pigmented skin lesion classification in
  images: review and future trends.
\newblock \emph{Neural Computing and Applications}, 29\penalty0 (3):\penalty0
  613--636, 2018.

\bibitem[Paszke et~al.(2017)Paszke, Gross, Chintala, Chanan, Yang, DeVito, Lin,
  Desmaison, Antiga, and Lerer]{paszke2017automatic}
A.~Paszke, S.~Gross, S.~Chintala, G.~Chanan, E.~Yang, Z.~DeVito, Z.~Lin,
  A.~Desmaison, L.~Antiga, and A.~Lerer.
\newblock Automatic differentiation in pytorch.
\newblock 2017.

\bibitem[Perez et~al.(2018)Perez, Vasconcelos, Avila, and Valle]{perez2018data}
F.~Perez, C.~Vasconcelos, S.~Avila, and E.~Valle.
\newblock Data augmentation for skin lesion analysis.
\newblock In \emph{OR 2.0 Context-Aware Operating Theaters, Computer Assisted
  Robotic Endoscopy, Clinical Image-Based Procedures, and Skin Image Analysis},
  pages 303--311. Springer, 2018.

\bibitem[Qian et~al.(2018)Qian, Liu, Jiang, Wang, Wang, Guan, and
  Sun]{qian2018two}
C.~Qian, T.~Liu, H.~Jiang, Z.~Wang, P.~Wang, M.~Guan, and B.~Sun.
\newblock A two-stage method for skin lesion analysis.
\newblock \emph{arXiv preprint arXiv:1809.03917}, 2018.

\bibitem[Ribeiro et~al.(2019)Ribeiro, Avila, and Valle]{ribeiro2019handling}
V.~Ribeiro, S.~Avila, and E.~Valle.
\newblock Handling inter-annotator agreement for automated skin lesion
  segmentation.
\newblock \emph{arXiv preprint arXiv:1906.02415}, 2019.

\bibitem[Rolnick et~al.(2017)Rolnick, Veit, Belongie, and
  Shavit]{rolnick2017deep}
D.~Rolnick, A.~Veit, S.~Belongie, and N.~Shavit.
\newblock Deep learning is robust to massive label noise.
\newblock \emph{arXiv preprint arXiv:1705.10694}, 2017.

\bibitem[Ronneberger et~al.(2015)Ronneberger, Fischer, and
  Brox]{ronneberger2015u}
O.~Ronneberger, P.~Fischer, and T.~Brox.
\newblock U-net: Convolutional networks for biomedical image segmentation.
\newblock pages 234--241, 2015.

\bibitem[Sivic and Zisserman(2006)]{sivic2006video}
J.~Sivic and A.~Zisserman.
\newblock Video google: Efficient visual search of videos.
\newblock In \emph{Toward category-level object recognition}, pages 127--144.
  Springer, 2006.

\bibitem[Tajbakhsh et~al.(2019)Tajbakhsh, Jeyaseelan, Li, Chiang, Wu, and
  Ding]{tajbakhsh2019embracing}
N.~Tajbakhsh, L.~Jeyaseelan, Q.~Li, J.~Chiang, Z.~Wu, and X.~Ding.
\newblock Embracing imperfect datasets: A review of deep learning solutions for
  medical image segmentation.
\newblock \emph{arXiv preprint arXiv:1908.10454}, 2019.

\bibitem[Tajbakhsh et~al.(2020)Tajbakhsh, Jeyaseelan, Li, Chiang, Wu, and
  Ding]{tajbakhsh2020embracing}
Nima Tajbakhsh, Laura Jeyaseelan, Qian Li, Jeffrey~N Chiang, Zhihao Wu, and
  Xiaowei Ding.
\newblock Embracing imperfect datasets: A review of deep learning solutions for
  medical image segmentation.
\newblock \emph{Medical Image Analysis}, page 101693, 2020.

\bibitem[Valle et~al.(2020)Valle, Fornaciali, Menegola, Tavares, Bittencourt,
  Li, and Avila]{valle2020data}
E.~Valle, M.~Fornaciali, A.~Menegola, J.~Tavares, F.~V. Bittencourt, L.~T. Li,
  and S.~Avila.
\newblock Data, depth, and design: Learning reliable models for skin lesion
  analysis.
\newblock \emph{Neurocomputing}, 383:\penalty0 303--313, 2020.

\bibitem[Van~der Walt et~al.(2014)Van~der Walt, Sch{\"o}nberger,
  Nunez-Iglesias, Boulogne, Warner, Yager, Gouillart, and Yu]{van2014scikit}
D.~Van~der Walt, J.~L. Sch{\"o}nberger, J.~Nunez-Iglesias, F.~Boulogne, J.~D.
  Warner, N.~Yager, E.~Gouillart, and T.~Yu.
\newblock scikit-image: image processing in python.
\newblock \emph{PeerJ}, 2:\penalty0 e453, 2014.

\bibitem[Xue et~al.(2018)Xue, Xu, and Huang]{xue2018adversarial}
Y.~Xue, T.~Xu, and X.~Huang.
\newblock Adversarial learning with multi-scale loss for skin lesion
  segmentation.
\newblock In \emph{International Symposium on Biomedical Imaging}, pages
  859--863, 2018.

\bibitem[Yuan(2017)]{yuan2017automatic}
Y.~Yuan.
\newblock Automatic skin lesion segmentation with fully
  convolutional-deconvolutional networks.
\newblock \emph{arXiv preprint arXiv:1703.05165}, 2017.

\bibitem[Zhao et~al.(2017)Zhao, Shi, Qi, Wang, and Jia]{zhao2017pyramid}
H.~Zhao, J.~Shi, X.~Qi, X.~Wang, and J.~Jia.
\newblock Pyramid scene parsing network.
\newblock In \emph{IEEE Conf. on Computer Vision and Pattern Recognition
  (CVPR)}, pages 2881--2890, 2017.

\end{thebibliography}
}

\end{document}